\documentclass{article}

\PassOptionsToPackage{numbers, compress}{natbib}

\usepackage[preprint]{neurips_2026}

\usepackage[utf8]{inputenc} 
\usepackage[T1]{fontenc}    
\usepackage{hyperref}       
\usepackage{url}            
\usepackage{booktabs}       
\usepackage{amsfonts}       
\usepackage{nicefrac}       
\usepackage{microtype}      
\usepackage{xcolor}         

\usepackage{amsmath}
\usepackage{amssymb}
\usepackage{todonotes}
\usepackage{enumitem}
\usepackage{array}
\usepackage{multirow}

\usepackage{subcaption} 
\usepackage{graphicx}
\usepackage[table]{xcolor}
\usepackage{diagbox}
\usepackage{bm}
\usepackage{algorithm}
\usepackage{algpseudocode}
\usepackage{amsmath}
\usepackage{float}
\usepackage{makecell}

\usepackage[none]{hyphenat}

\title{Learning to Perceive “Where”: Spatial Pretext Tasks for Robust Self-Supervised Learning}

\author{%
  Yang Shen\textsuperscript{1} \quad
  Yusen Cai\textsuperscript{1} \quad
  Weronika Hryniewska-Guzik\textsuperscript{1,2} \\[0.5em]
  \textbf{Qing Lin\textsuperscript{1,*}} \quad
  \textbf{Mengmi Zhang\textsuperscript{1,*}} \\[0.8em]
  \textsuperscript{1}Nanyang Technological University, Singapore \quad
  \textsuperscript{2}Warsaw University of Technology, Poland \\[0.3em]
  \textsuperscript{*}Co-corresponding authors \\[0.3em]
  \texttt{\{qing.lin,mengmi.zhang\}@ntu.edu.sg}
}

\begin{document}

\maketitle

\begin{abstract}
Existing self-supervised learning (SSL) methods primarily learn object-invariant representations but often neglect the spatial structure and relationships among object parts. To address this limitation, we introduce Spatial Prediction (SP), a spatially aware pretext regression task that predicts the relative position and scale between a pair of disentangled local views from the same image.
By modeling part-to-part relationships in a continuous geometric space, SP encourages representations to capture fine-grained spatial dependencies beyond invariant categorical semantics, thereby learning the compositional structure of visual scenes. SP is implemented as a decoupled plug-in and can be seamlessly integrated into diverse SSL frameworks.
Extensive experiments show consistent improvements across image recognition, fine-grained classification, semantic segmentation, and depth estimation, as well as substantial gains in out-of-distribution robustness for object recognition. To evaluate spatial reasoning, we introduce (1) a position and scale prediction task on image patch pairs and (2) a jigsaw understanding task requiring patch reordering and recognition after reconstruction. Strong performance on these tasks indicates improved spatial structure and geometric awareness. Overall, explicitly modeling spatial information provides an effective inductive bias for SSL, leading to more structured representations and better generalization. Code and models will be released.

\end{abstract}

\section{Introduction}
\label{sec:intro}
\begin{figure}[t]
  \centering
  \includegraphics[width=\linewidth]{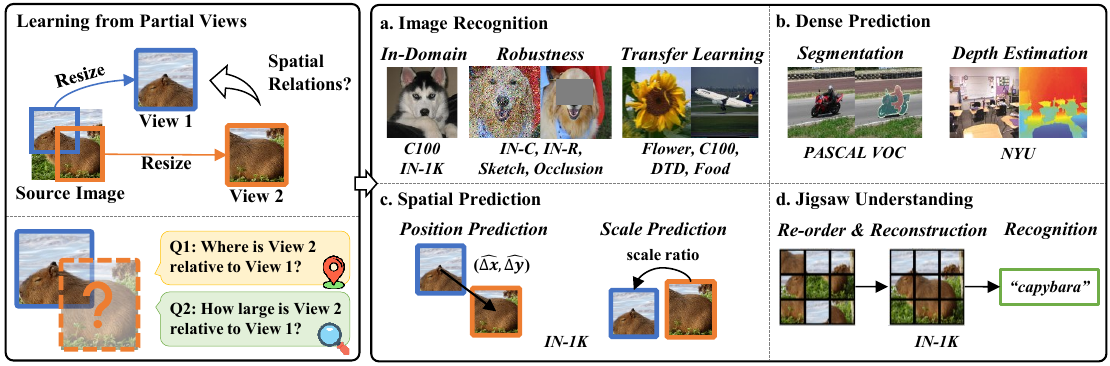}
  \caption{
    \textbf{Problem setting and task overview for visual representation learning from partial observations.} 
The left panel illustrates the problem setting: two cropped and resized views (orange and blue) are sampled from the same image, and the SSL model predicts the relative position and scale of the orange patch with respect to the blue reference.
The right panel summarizes the evaluation tasks for benchmarking SSL models. 
(a) Image recognition includes in-domain classification on C100~\cite{krizhevsky2009cifar100} and IN-1K~\cite{deng2009imagenet}, out-of-distribution robustness on IN-C~\cite{hendrycks2019imagenetc}, IN-R~\cite{hendrycks2021imagenetr}, Sketch~\cite{wang2019imagenetsketch}, and Occlusion, and cross-dataset transfer learning on Flowers~\cite{nilsback2008flower}, C100~\cite{krizhevsky2009cifar100}, DTD~\cite{cimpoi14dtd}, and Food~\cite{bossard14food101}. 
(b) Dense prediction tasks include semantic segmentation on PASCAL VOC~\cite{Everingham10pascalvoc} and depth estimation on NYU~\cite{Silberman2012nyuv2}. 
(c) Spatial prediction evaluates the ability of SSL models to estimate relative position and scale between two local views. 
(d) Jigsaw understanding evaluates patch reordering, reconstruction, and subsequent recognition. (c) and (d) correspond to our newly proposed benchmarks and tasks on IN-1K~\cite{deng2009imagenet}.
  }
  \label{fig:fig1}\vspace{-4mm}
\end{figure}

Humans exhibit a strong capacity for spatial perception, enabling them to infer the relative positions, scales, and arrangements of objects and their parts in complex scenes~\cite{biederman1987recognition, hummel1992dynamic}. 
Reasoning about part-to-part relationships is fundamental to understanding scene structure and visual composition, and underpins diverse real-world tasks, including object detection \cite{wu2023label,wang2024pose}, scene understanding \cite{zhang2020putting,bomatter2021pigs,liu2022reason,jia2025seeing,khandelwal2023adaptive}, instance segmentation \cite{wang2023object,han2024flow}, 3D reconstruction \cite{zhang2025peering}, depth estimation \cite{cai2025learning}, and visual navigation~\cite{piriyajitakonkij2024tta,yang2019spatialsense, liu2023visualspatial,zhang2018egocentric}. 
However, modern self-supervised learning (SSL) methods~\cite{Doersch2015contextprediction, assran2023ijepa, chen2021mocov3, caron2021dino, he2022mae, xie2022simmim, chen2020simclr, oquab2024dinov2, oord2018cpc} primarily emphasize semantic invariance or local reconstruction, while largely overlooking explicit modeling of spatial relationships between image regions. 
As a result, the geometric structure of visual scenes remains under-constrained.

This limitation stems from the absence of explicit supervision for spatial reasoning in the SSL literature. 
Invariance-based approaches, including contrastive and self-distillation methods~\cite{chen2021mocov3, caron2021dino, chen2021simsiam}, enforce consistency across augmented views, promoting robustness to transformations such as cropping and scaling. 
However, by treating these transformations as nuisances, they suppress spatial variation and reduce sensitivity to relative position and scale~\cite{o2020unsupervised, yun2022patch, pariza2024near, xie2021propagate}. 
In parallel, reconstruction-based methods~\cite{he2022mae, xie2022simmim, bao2022beit, assran2022masked} recover masked regions and capture fine-grained appearance statistics, but they operate at the patch level without spatial grounding. 
Consequently, learned representations exhibit limited sensitivity to part-to-part geometry~\cite{xie2023revealing}.

To close this gap, we propose \textbf{Spatial Prediction (SP)}, a spatially-aware pretext task that models geometric relationships between pairs of local regions via regression in continuous space. As illustrated in \textbf{Fig.~\ref{fig:fig1} (left)}, given two independently sampled views from the same image, the SSL model predicts their relative position and scale. By turning spatial relations into explicit regression objectives, SP encourages models to capture how local regions are organized within a scene, rather than discarding geometry as invariance or restricting representations to discrete grid-aligned feature tokens. SP is an architecture-agnostic plug-in that integrates seamlessly into existing SSL frameworks. It introduces an auxiliary spatial reasoning branch without modifying the original training objective, enhancing geometric sensitivity while preserving semantic robustness. The resulting representations are both invariant to appearance changes and sensitive to spatial variation, providing a more balanced inductive bias for visual representation learning.

We evaluate learned representations of SSL models from two complementary perspectives (\textbf{Fig.~\ref{fig:fig1}, right}). 
First, we assess the effectiveness of models trained with SP on standard computer vision tasks, including image recognition, fine-grained classification, out-of-distribution recognition under occlusion and corruption (\textbf{Fig.~\ref{fig:fig1}a}), as well as image segmentation and depth estimation (\textbf{Fig.~\ref{fig:fig1}b}). To further analyze spatial reasoning, we establish two diagnostic benchmarks: (1) a position and scale prediction task, which measures the ability to infer relative geometry between pairs of image patches (\textbf{Fig.~\ref{fig:fig1}c}); and (2) a jigsaw understanding task, which evaluates whether models can recover disrupted spatial layouts and perform recognition after reconstruction (\textbf{Fig.~\ref{fig:fig1}d}). Together, these tasks directly probe spatial awareness beyond conventional SSL benchmarks. Key contributions are highlighted: 

\noindent\textbf{1.} We introduce Spatial Prediction (SP), a spatially-aware pretext task that explicitly models geometric relationships between two local views by predicting their relative position and scale via regression. By formulating spatial relations as a direct training objective, SP goes beyond invariance- or reconstruction-driven SSL and encourages representations to capture fine-grained part-to-part dependencies and compositional scene structure.

\noindent\textbf{2.} We design SP as an architecture-agnostic plug-in that can be seamlessly integrated into diverse SSL frameworks, enabling spatial reasoning while preserving the semantic robustness of invariant representations.

\noindent\textbf{3.} We conduct extensive evaluations across five standard computer vision tasks, showing that SSL models trained with SP consistently outperform their counterparts. In addition, we introduce two spatial reasoning benchmarks: spatial prediction and jigsaw understanding. Experimental results provide direct evidence that SP improves spatial reasoning beyond standard benchmarks.

\section{Related Works on Self-Supervised Learning in Vision 
}\label{relatedwork}

SSL in vision seeks to learn visual representations from unlabeled data by designing pretext tasks that replace human annotations. 
Early vision-based approaches explored a variety of proxy objectives, including predicting relative patch positions~\cite{Doersch2015contextprediction}, patch re-ordering~\cite{noroozi2016jigsaw, misra2020selfsupervised}, image inpainting~\cite{pathak2016context}, colorization~\cite{zhang2016colorful}, and transformation prediction~\cite{gidaris2018rotationpred}.

\noindent\textbf{Reconstruction-based methods.}
With the advent of Vision Transformers (ViT)\cite{dosovitskiy2020vit}, masked image modeling (MIM) has emerged as a dominant paradigm. 
Methods such as MAE~\cite{he2022mae} and BEiT~\cite{bao2022beit}
learn representations by reconstructing masked or corrupted patches. Recent extensions, including latent-space prediction methods~\cite{baevski2022data2vec, assran2023ijepa}, further improve representation quality by predicting in feature space instead of pixel space, forming the JEPA family~\cite{lecun2022path}. 
Despite their success, recent studies~\cite{xie2023revealing, balestriero2024learning} indicate that these methods are often biased toward low-level texture reconstruction rather than holistic scene structure. 
Moreover, as noted in prior work~\cite{he2022mae, xie2022simmim}, these reconstruction-based methods operate on local patches without an explicit global coordinate system, limiting their ability to capture structured spatial dependencies.

\noindent\textbf{Discriminative SSL methods.}
A second line of work learns representations through discriminative objectives across different views of the same image. 
Beginning with instance discrimination~\cite{alexey2016discriminative, bojanowski2017unsupervised, wu2018unsupervised}, this family has evolved into contrastive learning methods~\cite{chen2020simclr, he2020moco, caron2020unsupervised, cai2025learning} such as MoCo\cite{he2020moco,chen2020mocov2,chen2021mocov3}, self-distillation approaches such as BYOL~\cite{Grill2020BYOL} and DINO~\cite{caron2021dino}, and clustering-based methods~\cite{caron2018deep, asano2019self, caron2020unsupervised}. 
These approaches achieve strong performance and transferability, particularly on ImageNet~\cite{deng2009imagenet}. 
Dense variants such as DenseCL~\cite{wang2021dense} extend contrastive learning to preserve local correspondences, while iBOT~\cite{zhou2022ibot} combines distillation with masked modeling to improve local feature learning. 
However, most of these methods still enforce view invariance without explicitly modeling spatial relationships between image regions, which can suppress fine-grained spatial structure~\cite{wang2017transitive, lee2026soft}.

\noindent\textbf{Spatial \& Position-based Pretext Tasks in SSL.}
Closely related to our work are methods that introduce spatial heuristics as training objectives. Early proxy tasks such as relative patch position prediction~\citep{Doersch2015contextprediction, zhai2022position} and jigsaw puzzle solving~\citep{noroozi2016jigsaw} inject spatial cues through handcrafted objectives. While effective, these approaches rely on heuristic designs and are not well aligned with modern large-scale SSL frameworks based on ViTs.
More recent methods incorporate spatial awareness via positional embeddings as auxiliary information. For example, I-JEPA~\citep{assran2023ijepa, chen2024context} predicts latent representations conditioned on positional embeddings, and DropPos~\citep{wang2023droppos} reconstructs masked positional coordinates from visible context. These methods improve spatial sensitivity but primarily treat position as conditioning rather than explicitly modeling their geometric relations.

Concurrently with our work, and most closely related, PART~\citep{ayoughi2025parts} also leverages relative position and scale prediction via regression for spatial reasoning. 
However, PART formulates this as a standalone pretext objective. In contrast, our SP is designed as a plug-in regularizer that complements existing SSL objectives, preserving semantic invariance while introducing explicit geometric supervision. 
Importantly, our SP introduces a rejection sampling strategy to curate pairs of local views with multiple constraints, ensuring that models learn from balanced spatial distributions, non-trivial yet informative relationships, and sufficiently informative views. This design promotes stable spatial supervision and avoids degenerate cases caused by excessive overlap, biased target distributions, or overly small patches with limited semantic content. In contrast, PART does not explicitly control the sampling distribution of view pairs.
Empirically, compared to PART, SP is evaluated across a broader range of downstream tasks and dedicated spatial reasoning benchmarks, demonstrating consistent improvements in both robust semantic representation and spatial reasoning ability, whereas evaluations in PART are rather limited.
\section{Spatial Prediction (SP): A Spatially-Aware Pretext Task}
\label{sec: sp_method}

\begin{figure}[t]
  \centering
  \includegraphics[width=\linewidth]{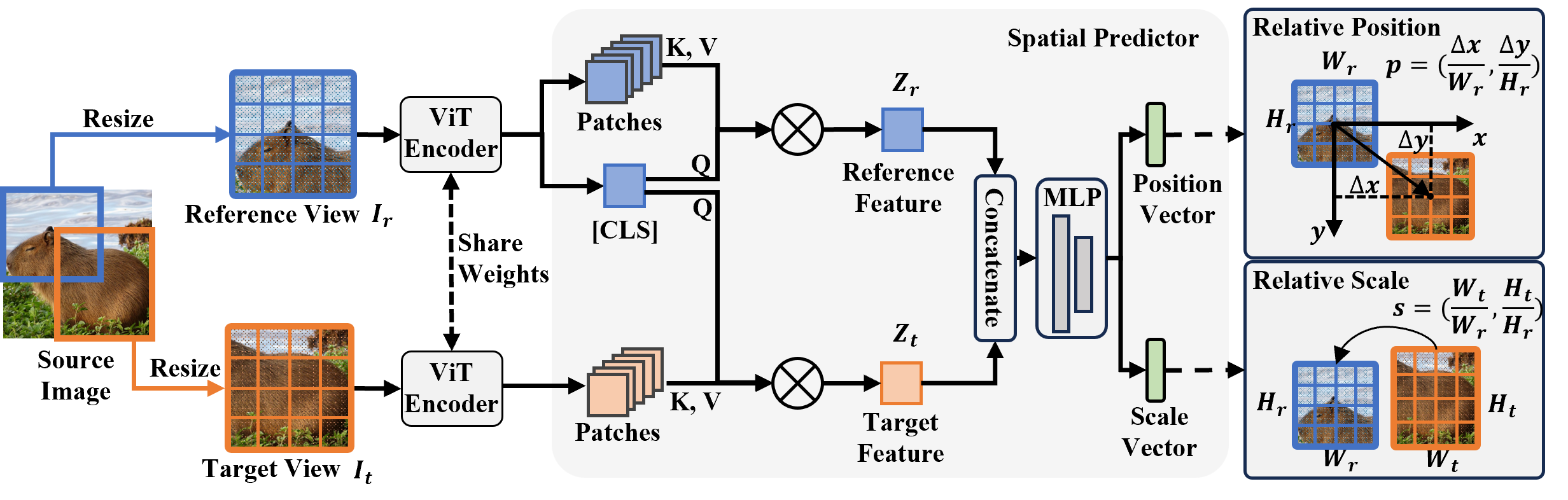}
  \caption{
   \textbf{Overview of Spatial Prediction (SP).}
Given a reference view $I_r$ and a target view $I_t$ sampled from the same image, both views are encoded by a shared Vision Transformer (ViT)~\cite{dosovitskiy2020vit}. 
The class token ([CLS]) of the reference view serves as the query ($Q$), while patch tokens from both reference and target views act as keys ($K$) and values ($V$) to compute a cross-attention-like interaction ($\otimes$), producing reference and target features. 
This operation is parameter-free, as it directly uses token embeddings without additional projections for $Q$, $K$, and $V$.
The resulting features $\mathbf{Z}_r$ and $\mathbf{Z}_t$ are concatenated and fed into a two-layer MLP to regress the relative position and scale, supervised using $\ell_2$ loss against ground-truth targets. 
As illustrated in the right panels, the ground-truth spatial relationship is defined by the relative offset $(\Delta x, \Delta y)$ from the reference view to the target view, along with the relative scale normalized by the reference view dimensions $(H_r, W_r)$, where $(H_r, W_r)$ and $(H_t, W_t)$ denote the height and width of the reference and target views, respectively.}
  \label{fig:fig2}
\end{figure}

We propose a plug-and-play pretext task, termed \textbf{Spatial Prediction (SP)}, that augments SSL with explicit geometric supervision during pre-training (\textbf{Fig.~\ref{fig:fig2}}). It can be seamlessly integrated into diverse SSL frameworks without modifying their model architectures. 
SP leverages Vision Transformer (ViT)~\cite{dosovitskiy2020vit} feature tokens to model spatial dependencies.
At a high level, SP shares the encoder $\psi(\cdot)$ with the original SSL objective and introduces an additional spatial prediction branch. 
Both objectives are jointly optimized during pre-training, enabling the model to learn complementary semantic and geometric representations.
Given an input image $\mathbf{I} \in \mathbb{R}^{H\times W \times 3}$, where $H$ and $W$ denote the height and width, the encoder $\psi(\cdot)$ outputs a sequence of patch tokens $\mathbf{Z} \in \mathbb{R}^{L\times D}$ and a classification token $\mathbf{z} \in \mathbb{R}^{1\times D}$, with $L$ denoting the number of tokens and $D$ the feature dimension.
SP operates on $\mathbf{Z}$ and $\mathbf{z}$, as described below. 



\subsection{Curating Pairs of Augmented Local Views for Pre-Texting Tasks}
\label{sec:spatial_formulation}

To train SSL models with SP, we construct pairs of augmented local views from the same image. Given a source image $\mathbf{I}$, we first apply standard augmentations (e.g., random horizontal flipping), followed by rejection sampling to obtain two local views. The sampling is guided by three criteria: (i) views should be sufficiently local to capture partial object information, but not so small that they lose semantic content; (ii) the relative position and scale targets are sampled to be approximately uniform, avoiding skewed or long-tailed distributions that could bias regression toward trivial solutions; and (iii) a non-trivial spatial displacement is enforced to prevent large overlaps that would make the prediction task overly easy due to shortcut cues. We validate the distributions of ground truth relative position and scale after rejection sampling in \textbf{Fig.\ref{fig:figS2}}.





To further mitigate shortcut solutions, we resize both sampled local views to a fixed size and apply independent color augmentations. We avoid extra spatial transformations (e.g., rotation or further cropping) that would distort the ground-truth relative positions and scales. These design choices discourage the SSL models from over-reliance on low-level cues such as color or template matching, and instead promote semantic correspondence for spatial reasoning.

Without loss of generality, we denote the first local view as the reference view $I_r$ and the second as the target view $I_t$, with corresponding heights and widths $(H_r, W_r)$ and $(H_t, W_t)$, respectively. Since $I_r$ and $I_t$ may differ in size and location, we define the ground-truth relative position $p$ and scale $s$ normalized by the reference view dimensions $(H_r, W_r)$ (see \textbf{Fig.~\ref{fig:fig2}}). 
This normalization is a key design choice, as it makes the geometric ground truth scale-invariant and independent of the absolute image size of $I_r$, allowing SSL models to focus on relative spatial structure.

\subsection{Spatial Representation Learning}
\label{sec:representation_learning}

The encoder $\psi(\cdot)$ takes the reference and target views $I_r$ and $I_t$ as inputs and produces their patch tokens and classification tokens, where $Z_r, Z_t \in \mathbb{R}^{L\times D}$ and $z_r, z_t \in \mathbb{R}^{1\times D}$:
\begin{equation}
[Z_r, z_r] = \psi(I_r), \quad [Z_t, z_t] = \psi(I_t).
\end{equation}

Motivated by human spatial reasoning, we formulate the task as reference-conditioned spatial inference, where a reference view provides an anchor for estimating the relative position and scale of a target view. The model first establishes a reference coordinate system from $I_r$ and then performs spatial reasoning over $I_t$ conditioned on this reference.
Thus, we introduce a cross-attention-based mechanism to model this interaction. Specifically, $z_r$ serves as a query, while $Z_r$ and $Z_t$ serve as keys and values for retrieving reference-aware and target-aware features:
\begin{equation}
h_r = \text{Softmax}\big(\mathrm{norm}(z_r) Z_r^\top\big) Z_r, \quad
h_t = \text{Softmax}\big(\mathrm{norm}(z_r) Z_t^\top\big) Z_t,
\end{equation}
where $\top$ denotes matrix transpose and $\mathrm{norm}(\cdot)$ is layer normalization for training stability. Unlike standard cross-attention, we directly use patch tokens and classification tokens without additional linear projections for queries, keys, and values. This design enforces stronger supervision on token interactions for spatial reasoning while avoiding additional learnable parameters.

The retrieved features are then concatenated to form a joint representation. A spatial predictor $\theta(\cdot)$, implemented as a two-layer MLP with hidden dimension 384 and ReLU activation, regresses the spatial outputs:
\begin{equation}
[\hat{p}, \hat{s}] = \theta([h_r; h_t]),
\end{equation}
where $\hat{p} \in \mathbb{R}^2$ and $\hat{s} \in \mathbb{R}^2$ denote the predicted relative spatial displacement and scale, respectively.

\subsection{Training Objectives and Implementation Details}
\label{sec:objectives}
The proposed SP objective is introduced as an auxiliary loss in addition to the base SSL objective, denoted as $\mathcal{L}_{\text{base}}$ (e.g., contrastive loss for MoCo v3~\citep{chen2021mocov3}, distillation loss for DINO~\cite{caron2021dino}, or reconstruction loss for MAE~\cite{he2022mae}). Detailed schematics of SP integration into each SSL framework are provided in \textbf{Fig.~\ref{fig:figS1}}.

The spatial predictor is trained to regress the relative spatial configuration between two views using $\ell_2$ losses:
\begin{equation}
\mathcal{L}_{\text{total}} = \mathcal{L}_{\text{base}} + \mathcal{L}_{\text{SP}}, \quad \text{where} \quad
\mathcal{L}_{\text{SP}} =
\lambda_p \|\hat{\mathbf{p}} - \mathbf{p}\|_2^2 +
\lambda_s \|\hat{\mathbf{s}} - \mathbf{s}\|_2^2,
\end{equation}
$\lambda_p = 0.1$ and $\lambda_s = 0.1$ balance the two regression terms. See \textbf{Tab.~\ref{tab:ablation_lambda}} for hyperparameter analysis.

We pre-train all SSL models with SP from scratch. All experiments are conducted on two NVIDIA RTX A6000 GPUs with a batch size of 256. Optimization uses AdamW with a cosine learning-rate schedule. All hyperparameters follow the default configurations of the original SSL implementations without SP. See \textbf{Tab.~\ref{tab:hyperparams_final}} for a summary of pre-training hyperparameters for all SSL models.

\section{Experiments}
\label{sec:Experiments}
We introduce a comprehensive evaluation suite covering 7 downstream tasks, 11 datasets, 6 SSL models, 2 backbones, and 7 metrics.

\subsection{Datasets and Metrics}
\label{sec: datasets and metrics}

\noindent \textbf{Datasets for pre-training SSL models:}
\textbf{CIFAR-100 (C100)}~\cite{krizhevsky2009cifar100}, a small-scale dataset with 60K images of size $32\times32$ across 100 classes, and \textbf{ImageNet-1K (IN-1K)}~\cite{deng2009imagenet}, a large-scale dataset with 1.28M images of size $224\times224$ across 1,000 classes. 
For in-domain image classification, we evaluate on the corresponding test sets. 
For all subsequent tasks, we use SSL backbones pretrained on IN-1K.
We report Top-1 classification accuracy (Acc in \%) of SSL models. 

\noindent \textbf{Datasets for robustness tests in image classification:}
We evaluate SSL model robustness using several ImageNet-derived benchmarks. 
\textbf{ImageNet-C (IN-C)}~\cite{hendrycks2019imagenetc} measures robustness to common corruptions (e.g., noise, blur, weather) across multiple severity levels. 
\textbf{ImageNet-R (IN-R)}~\cite{hendrycks2021imagenetr} and \textbf{ImageNet-Sketch (Skt)}~\cite{wang2019imagenetsketch} assess robustness to domain and appearance shifts, including artistic renditions and sketch-based images that emphasize shape over texture and color. 
\textbf{ImageNet-Occlusion (Occ)} is our synthetic benchmark, where we systematically apply occlusions by overlaying opaque rectangular masks at random locations with varying coverage ratios of 0.1, enabling controlled evaluation of robustness to partial visibility and spatial reasoning under missing visual information.
We report Top-1 classification accuracy (Acc. in \%) of SSL models across all datasets, except IN-C, where we report mean Corruption Error (mCE), computed as the average normalized error over all corruption types and severity levels.

\noindent \textbf{Datasets for transfer learning in fine-grained classification:}
We evaluate transfer learning ability of SSL models on diverse fine-grained datasets. 
\textbf{Flowers-102 (Flower)}~\cite{nilsback2008flower} contains 102 flower categories with subtle inter-class differences. 
\textbf{DTD}~\cite{cimpoi14dtd} focuses on texture recognition across 47 categories of materials and surface patterns. 
\textbf{Food-101 (Food)}~\cite{bossard14food101} includes 101 food categories with high visual variability. We report Top-1 classification accuracy (Acc in \%) of SSL models.

\noindent \textbf{Dataset for transfer learning in semantic segmentation:}
We evaluate semantic segmentation performance of SSL models. 
\textbf{PASCAL VOC (VOC)}~\cite{Everingham10pascalvoc} is used for semantic segmentation, with pixel-level annotations on images of size $512\times512$ across 21 categories.
We report mean Intersection over Union (mIoU), computed as the average intersection-over-union between predicted and ground-truth segmentation masks across classes.

\noindent \textbf{Dataset for transfer learning in depth estimation:}
We evaluate depth estimation performance of SSL models.
\textbf{NYU v2 (NYU)}~\cite{Silberman2012nyuv2} is used for depth estimation, providing aligned RGB images and depth maps for indoor scenes.
We report root mean squared error (RMSE), computed as the square root of the mean squared difference between predicted and ground-truth depth values.

\noindent \textbf{Dataset for spatial prediction:}
We use images from the IN-1K test set and sample pairs of local patches from the same image following the pre-training protocol (\textbf{Sec.\ref{sec: sp_method}}). Given a reference patch, the model is required to predict the relative position and scale of the target patch. We report L2 distance between the predicted and ground-truth relative position and scale.

\noindent \textbf{Dataset for jigsaw understanding:}
We use images from the IN-1K test set, partition each image into a $3\times3$ grid, and shuffle the patches using a predefined set of 1,000 permutations. 
Models are required to predict the permutation index (1,000-way classification), reconstruct the image, and perform recognition on the restored input.
We report Top-1 accuracy (Acc in \%), measuring (i) permutation classification accuracy and (ii) object recognition accuracy on reconstructed images.





\subsection{Baselines, Backbones, and Experimental Protocols}

We evaluate three representative SSL frameworks: \textbf{MoCo v3}~\citep{chen2021mocov3}, \textbf{DINO}~\citep{caron2021dino}, and \textbf{MAE}~\citep{he2022mae}, covering contrastive learning, self-distillation, and reconstruction-based pretraining. We follow their original data augmentation, pre-training, linear probing, and fine-tuning protocols~\citep{chen2021mocov3, caron2021dino, he2022mae}.
For SSL methods integrated with our SP module, we retain their default pre-training, linear probing, and fine-tuning configurations, consistent with their standard counterparts without SP~\citep{he2022mae, caron2021dino, oquab2024dinov2, assran2023ijepa}. In linear probing, the backbone is frozen and only a linear classifier is trained end-to-end. The learning rate is selected via grid search based on validation performance.

Under a fixed computational budget, we use ViT-S/16~\citep{dosovitskiy2020vit} for MoCo v3 and DINO, and ViT-B/16~\citep{dosovitskiy2020vit} for MAE. We note that using different backbone sizes does not affect fairness of comparison, as we focus on within-model comparisons with and without SP.

For in-domain image classification, fine-grained transfer learning, and depth estimation, we report linear probing results.
For robustness evaluation, we directly apply the same linear probes to the target datasets without additional fine-tuning.
For semantic segmentation, we report fine-tuning results.

For the spatial prediction task, we use the same cross-attention-based mechanism as in spatial representation learning to predict the relative position and scale between two local views (\textbf{Sec.~\ref{sec:representation_learning}}).

For the jigsaw understanding task, we independently encode 9 shuffled image patches using the SSL backbone and extract their classification tokens. These classification tokens serve as queries that attend over all patch tokens (keys and values) from all patches via a cross-attention mechanism, as described in \textbf{Sec.~\ref{fig:figS4}}. The resulting outputs from the 9 cross-attention operations are concatenated and passed through a 2-layer MLP to predict the permutation index. For object recognition on reconstructed images, we reuse the same linear probes trained for in-domain classification without further fine-tuning.

\section{Results}
\label{sec: result}

\begin{table}[t]
\centering
\caption{\textbf{Performance comparison across all downstream tasks except spatial reasoning.}
Each column corresponds to a task, together with its dataset and evaluation metric. Rows are grouped in pairs for SSL models with and without SP, where each pair shares the same backbone.
All reported Acc values denote Top-1 classification accuracy (\%). For IN-C, VOC, and NYU, we report mCE$\downarrow$, mIoU$\uparrow$, and RMSE (scaled by $\times 10^2$)$\downarrow$, respectively. See \textbf{Sec.~\ref{sec:Experiments}} for experimental details. Each experiment is repeated three times, and we report the mean performance with standard deviations in ($\pm$ std). Best results are shown in bold.
}
\label{tab:semantic_table}
\resizebox{\textwidth}{!}{%
\setlength{\tabcolsep}{4pt}
\begin{tabular}{lcccccccccccc}
\toprule
\textbf{} & \multicolumn{10}{c}{Image Recognition} & \multicolumn{2}{c}{Dense Prediction} \\ 
\cmidrule(lr){2-11} \cmidrule(lr){12-13}
\textbf{} & \multicolumn{2}{c}{a. In-Domain} & \multicolumn{4}{c}{b. Robustness} & \multicolumn{4}{c}{c. Transfer Learning} & d. Seg. & e. Depth \\ 
\cmidrule(lr){2-3} \cmidrule(lr){4-7} \cmidrule(lr){8-11} \cmidrule(lr){12-12} \cmidrule(lr){13-13}
& C100* & IN-1K & IN-C & IN-R & Skt & Occ & Flw & C100 & DTD & Food & VOC & NYU \\

& \cite{krizhevsky2009cifar100} & \cite{deng2009imagenet} & \cite{hendrycks2019imagenetc} & \cite{hendrycks2021imagenetr} & \cite{wang2019imagenetsketch} &  & \cite{nilsback2008flower} & \cite{krizhevsky2009cifar100} & \cite{cimpoi14dtd} & \cite{bossard14food101} & \cite{Everingham10pascalvoc} & \cite{Silberman2012nyuv2} \\

& Acc↑ & Acc↑ & mCE↓ & Acc↑ & Acc↑ & Acc↑ & Acc↑ & Acc↑ & Acc↑ & Acc↑ & mIoU↑ & RMSE↓\\ 
\midrule

\raisebox{0.6ex}{\shortstack[c]{MAE\cite{he2022mae}}}
& \shortstack[c]{39.7\\ \footnotesize $\pm$ 0.5}
& \shortstack[c]{44.6\\ \footnotesize $\pm$ 0.6}
& \shortstack[c]{108.5\\ \footnotesize $\pm$ 0.3}
& \shortstack[c]{7.2\\ \footnotesize $\pm$ 0.3}
& \shortstack[c]{5.5\\ \footnotesize $\pm$ 0.3}
& \shortstack[c]{36.7\\ \footnotesize $\pm$ 0.3}
& \shortstack[c]{72.7\\ \footnotesize $\pm$ 0.2}
& \shortstack[c]{65.6\\ \footnotesize $\pm$ 0.1}
& \shortstack[c]{55.5\\ \footnotesize $\pm$ 0.2}
& \shortstack[c]{56.8\\ \footnotesize $\pm$ 0.2}
& \shortstack[c]{51.0\\ \footnotesize $\pm$ 0.1}
& \shortstack[c]{58.2\\ \footnotesize $\pm$ 0.2} \\
\addlinespace[0.3em]

\rowcolor{black!4}
\raisebox{0.6ex}{\shortstack[c]{\quad +SP}}
& \shortstack[c]{\textbf{43.5}\\ \footnotesize \textbf{$\pm$ 0.4}}
& \shortstack[c]{\textbf{52.6}\\ \footnotesize \textbf{$\pm$ 0.4}}
& \shortstack[c]{\textbf{104.8}\\ \footnotesize \textbf{$\pm$ 0.2}}
& \shortstack[c]{\textbf{9.2}\\ \footnotesize \textbf{$\pm$ 0.5}}
& \shortstack[c]{\textbf{7.6}\\ \footnotesize \textbf{$\pm$ 0.4}}
& \shortstack[c]{\textbf{41.6}\\ \footnotesize \textbf{$\pm$ 0.9}}
& \shortstack[c]{\textbf{80.1}\\ \footnotesize \textbf{$\pm$ 0.2}}
& \shortstack[c]{\textbf{67.6}\\ \footnotesize \textbf{$\pm$ 0.2}}
& \shortstack[c]{\textbf{56.5}\\ \footnotesize \textbf{$\pm$ 0.2}}
& \shortstack[c]{\textbf{60.7}\\ \footnotesize \textbf{$\pm$ 0.2}}
& \shortstack[c]{\textbf{51.2}\\ \footnotesize \textbf{$\pm$ 0.2}}
& \shortstack[c]{\textbf{57.3}\\ \footnotesize \textbf{$\pm$ 0.3}} \\
 \midrule

\raisebox{0.6ex}{\shortstack[c]{MoCo v3\cite{chen2021mocov3}}}
& \shortstack[c]{62.1\\ \footnotesize $\pm$ 0.3}
& \shortstack[c]{61.1\\ \footnotesize $\pm$ 0.6}
& \shortstack[c]{85.6\\ \footnotesize $\pm$ 0.2}
& \shortstack[c]{13.9\\ \footnotesize $\pm$ 0.2}
& \shortstack[c]{13.6\\ \footnotesize $\pm$ 0.1}
& \shortstack[c]{57.3\\ \footnotesize $\pm$ 0.3}
& \shortstack[c]{63.7\\ \footnotesize $\pm$ 0.4}
& \shortstack[c]{62.2\\ \footnotesize $\pm$ 0.1}
& \shortstack[c]{57.2\\ \footnotesize $\pm$ 0.3}
& \shortstack[c]{58.5\\ \footnotesize $\pm$ 0.1}
& \shortstack[c]{58.5\\ \footnotesize $\pm$ 0.1}
& \shortstack[c]{60.1\\ \footnotesize $\pm$ 0.7} \\
\addlinespace[0.3em]

\rowcolor{black!4}
\raisebox{0.6ex}{\shortstack[c]{\quad +SP}}
& \shortstack[c]{\textbf{64.6}\\ \footnotesize \textbf{$\pm$ 0.3}}
& \shortstack[c]{\textbf{66.1}\\ \footnotesize \textbf{$\pm$ 0.6}}
& \shortstack[c]{\textbf{81.4}\\ \footnotesize \textbf{$\pm$ 0.2}}
& \shortstack[c]{\textbf{16.7}\\ \footnotesize \textbf{$\pm$ 0.1}}
& \shortstack[c]{\textbf{15.9}\\ \footnotesize \textbf{$\pm$ 0.1}}
& \shortstack[c]{\textbf{60.5}\\ \footnotesize \textbf{$\pm$ 0.9}}
& \shortstack[c]{\textbf{88.7}\\ \footnotesize \textbf{$\pm$ 1.1}}
& \shortstack[c]{\textbf{77.2}\\ \footnotesize \textbf{$\pm$ 0.1}}
& \shortstack[c]{\textbf{66.7}\\ \footnotesize \textbf{$\pm$ 0.3}}
& \shortstack[c]{\textbf{72.8}\\ \footnotesize \textbf{$\pm$ 0.2}}
& \shortstack[c]{\textbf{60.2}\\ \footnotesize \textbf{$\pm$ 0.1}}
& \shortstack[c]{\textbf{59.9}\\ \footnotesize \textbf{$\pm$ 0.6}} \\
\midrule

\raisebox{0.6ex}{\shortstack[c]{DINO\cite{caron2021dino}}}
& \shortstack[c]{50.5\\ \footnotesize $\pm$ 0.2}
& \shortstack[c]{72.8\\ \footnotesize $\pm$ 0.2}
& \shortstack[c]{\textbf{86.3}\\ \footnotesize \textbf{$\pm$ 0.2}}
& \shortstack[c]{\textbf{15.7}\\ \footnotesize \textbf{$\pm$ 0.1}}
& \shortstack[c]{\textbf{14.6}\\ \footnotesize \textbf{$\pm$ 0.1}}
& \shortstack[c]{60.1\\ \footnotesize $\pm$ 0.3}
& \shortstack[c]{86.4\\ \footnotesize $\pm$ 0.5}
& \shortstack[c]{75.8\\ \footnotesize $\pm$ 0.1}
& \shortstack[c]{\textbf{66.5}\\ \footnotesize \textbf{$\pm$ 0.2}}
& \shortstack[c]{\textbf{70.5}\\ \footnotesize \textbf{$\pm$ 0.2}}
& \shortstack[c]{46.4\\ \footnotesize $\pm$ 0.1}
& \shortstack[c]{57.3\\ \footnotesize $\pm$ 0.9} \\
\addlinespace[0.3em]

\rowcolor{black!4}
\raisebox{0.6ex}{\shortstack[c]{\quad +SP}}
& \shortstack[c]{\textbf{53.0}\\ \footnotesize \textbf{$\pm$ 0.4}}
& \shortstack[c]{\textbf{73.0}\\ \footnotesize \textbf{$\pm$ 0.2}}
& \shortstack[c]{88.5\\ \footnotesize $\pm$ 0.3}
& \shortstack[c]{\textbf{15.7}\\ \footnotesize \textbf{$\pm$ 0.1}}
& \shortstack[c]{\textbf{14.6}\\ \footnotesize \textbf{$\pm$ 0.1}}
& \shortstack[c]{\textbf{60.5}\\ \footnotesize \textbf{$\pm$ 0.7}}
& \shortstack[c]{\textbf{87.3}\\ \footnotesize \textbf{$\pm$ 0.2}}
& \shortstack[c]{\textbf{76.1}\\ \footnotesize \textbf{$\pm$ 0.2}}
& \shortstack[c]{65.7\\ \footnotesize $\pm$ 0.1}
& \shortstack[c]{70.3\\ \footnotesize $\pm$ 0.2}
& \shortstack[c]{\textbf{47.2}\\ \footnotesize \textbf{$\pm$ 0.1}}
& \shortstack[c]{\textbf{55.8}\\ \footnotesize \textbf{$\pm$ 0.3}} \\
\bottomrule
\end{tabular}%
}\vspace{-4mm}
\end{table}

\subsection{SP Excels at Spatial Reasoning While Preserving Robust Semantic Representations.}
\label{sec: result_semantic}

\noindent \textbf{SP does not impair, but instead enhances semantic representation learning.}
As shown in \textbf{Table~\ref{tab:semantic_table}a}, SSL models with SP achieve consistent gains over their baselines on both CIFAR-100 and IN-1K for in-domain image classification. The improvements are particularly pronounced for MAE and MoCo v3, while even the strong DINO benefits from SP. These results indicate that explicit spatial supervision does not hinder semantic representation learning; instead, it enhances the semantic quality of learned representations.

\begin{figure}
    \centering
    \includegraphics[width=1\linewidth]{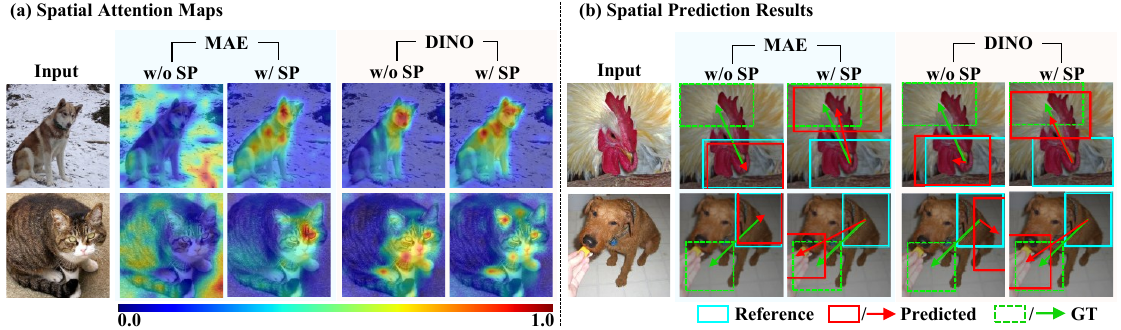}
    \caption{
    \textbf{Example visualizations of learned representations by SSL models with and without SP.}
\textbf{(a) Qualitative comparison of spatial attention maps.}
Following~\cite{caron2021dino}, we visualize attention by computing the attention weights between the classification token $z$ and all patch tokens $Z$, reshaping them into a 2D spatial map, and upsampling to the input image size.
Each row corresponds to an input image, and columns show attention maps overlaid on the images for MAE and DINO, with and without SP. Heatmaps are normalized to [0,1]; see the colorbar for scale.
\textbf{(b) Qualitative comparison of spatial reasoning results.}
Predicted relative positions and scales between two local views are visualized on the original image.
Each row corresponds to an input image, and columns show results for MAE and DINO, with and without SP. Blue boxes denote reference views, red boxes denote predicted positions and scales, and green dashed boxes denote ground truth; arrows indicate predicted (red) and ground-truth (green) displacement vectors. See \textbf{Fig.~\ref{fig:figS3}} for examples on MoCo v3.
    }\vspace{-2mm}
    \label{fig:fig3}
\end{figure}

To further analyze model behavior in in-domain classification, we visualize 2D attention maps on example images in \textbf{Fig.~\ref{fig:fig3}}, illustrating where SSL models attend during classification. Without SP, attention maps (e.g., a sitting dog in Row 1) are often diffuse and partially influenced by background context. In contrast, models with SP focus more on semantically meaningful object regions. While DINO already highlights the dog face, DINO with SP yields sharper localization and more comprehensive coverage of the dog body. 
This behavior suggests that SP encourages object-centric representations, which further improves the quality of learned semantic features.


\noindent \textbf{SP improves robustness to noise, occlusion and corruptions and reduces bias towards textures.}
We evaluate robustness on IN-C, IN-R, Skt, and Occ without fine-tuning. As shown in \textbf{Tab.~\ref{tab:semantic_table}b}, SP consistently improves performance for MAE and MoCo v3 across all benchmarks.
Gains are particularly notable on Skt and IN-R, where texture cues are suppressed and shape information dominates; e.g., MoCo v3 + SP improves by +2.3\% 
on Skt and +2.8\% on IN-R. This aligns with prior findings that SSL models without SP tend to over-rely on texture cues~\cite{cai2025learning}. On IN-C, SP reduces mCE across all backbones (e.g., 108.5$\rightarrow$104.8 for MAE and 85.6$\rightarrow$81.4 for MoCo v3), indicating improved robustness to noise and blur.
These results suggest that SP introduces structural inductive bias, encouraging reliance on global object layout when local cues are degraded.



\noindent \textbf{SP enhances feature transferability, especially for fine-grained recognition.}
We evaluate transfer learning on Flw, Food, DTD, and C100. As shown in \textbf{Tab.~\ref{tab:semantic_table}c}, SP consistently improves performance across most datasets. Gains are particularly large on fine-grained tasks; e.g., MoCo v3 + SP improves by 25\% 
on Flw, highlighting the benefit of modeling part-level spatial relationships. Such relational representations are well suited for fine-grained recognition, where subtle part-to-whole configurations are critical.

\noindent \textbf{SP provides spatially grounded features for semantic segmentation.}
We evaluate semantic segmentation on VOC. As shown in \textbf{Tab.~\ref{tab:semantic_table}d}, SP consistently improves mIoU across all SSL methods.
These results suggest that SP enhances spatial grounding in learned features, benefiting pixel-level prediction. By preserving localized semantic structures rather than collapsing representations into a global descriptor, SP enables better exploitation of spatial information during fine-tuning, leading to improved object boundaries and precise localization.

\noindent \textbf{SP enables representations that capture geometric structure for depth estimation.}
We evaluate depth estimation on NYU. As shown in \textbf{Tab.~\ref{tab:semantic_table}e}, SP consistently reduces RMSE across all backbones, with the largest improvement for DINO (-1.5 
RMSE). These results suggest that SP promotes representations that capture 3D geometric information. By explicitly modeling spatial relationships during pre-training, SP encourages the use of geometric cues such as depth ordering and surface continuity for spatial inference.

\noindent \textbf{SP provides stronger spatial supervision than implicit positional embeddings in standard SSL.}
We evaluate spatial prediction via relative position and scale estimation. As shown in \textbf{Tab.~\ref{tab:tab2}a--b}, incorporating SP consistently reduces L2 errors for both position and scale across all SSL backbones. These results indicate that SP provides explicit geometric supervision, enabling representations to encode precise spatial relationships beyond implicit positional embeddings used in standard SSL.

Visualization results in \textbf{Fig.~\ref{fig:fig3}b} further support this observation. Standard SSL models often exhibit noticeable errors in both the direction and magnitude of relative position offsets, reflected by misaligned displacement vectors and bounding boxes. In contrast, models with SP produce predictions whose spatial predictions (red boxes) closely match the ground truth (green dashed boxes), with displacement vectors well aligned in both direction and magnitude.



\begin{table*}[t]
    \centering
    \begin{minipage}[t]{0.49\textwidth}
        \centering
        \caption{\textbf{Performance comparison in spatial prediction and jigsaw understanding tasks.}
        Columns are grouped into spatial prediction (position and scale; evaluated by L2 distance $\downarrow$) and jigsaw understanding (permutation order prediction and recognition on reconstructed images; evaluated by Top-1 accuracy $\uparrow$ (\%)).
        Rows are grouped in pairs, each corresponding to a SSL method with and w/o our SP.
        Best in bold.}
        \label{tab:tab2}
        \resizebox{0.96\textwidth}{!}{
        \setlength{\tabcolsep}{4pt}
        \begin{tabular}{l cc cc}
        \toprule
         & \multicolumn{2}{c}{Spatial Prediction} & \multicolumn{2}{c}{Jigsaw Understanding} \\
        \cmidrule(lr){2-3} \cmidrule(lr){4-5}
        \textbf{} & a. Position & b. Scale & c. Order & d. Recog. \\
        \midrule
        
        MAE\cite{he2022mae} & 0.92 & 0.39 & 77.95 & 39.19 \\
        
        \rowcolor{black!4}
        \quad +SP & \textbf{0.61} & \textbf{0.35} & \textbf{98.58} & \textbf{48.21} \\
        \addlinespace[0.5em]
        
        MoCo v3\cite{chen2021mocov3} & 1.47 & 0.45 & 69.87 & 57.09 \\

        \rowcolor{black!4}
        \quad +SP & \textbf{1.16} & \textbf{0.41} & \textbf{90.24} & \textbf{64.23} \\
        \addlinespace[0.5em]
        
        DINO\cite{caron2021dino} & 1.32 & 0.43 & 88.45 & 63.45 \\

        \rowcolor{black!4}
        \quad +SP & \textbf{1.20} & \textbf{0.42} & \textbf{96.17} & \textbf{64.56} \\
        \bottomrule
        \end{tabular}%
        }
        \vspace{1pt}
    \end{minipage}
    \hfill
    \begin{minipage}[t]{0.49\textwidth}
        \centering
        \caption{
        \textbf{Analysis of SP design choices.}
L2 regression supervises relative position (Col1) and scale (Col2). We compare cross-attention mechanisms with (PAttn, Col3) and without (FAttn, Col4) linear projections for queries, keys, and values, where checkmarks (\checkmark) indicate their presence. Experiments are conducted using MoCo v3 on C100 for in-domain image classification. Performance is reported as Top-1 accuracy (\%)$\uparrow$, with the best results in bold.
}
        \label{tab:ablation_checkmark_optimized}
        \resizebox{\textwidth}{!}{
        \setlength{\tabcolsep}{6pt}
        \begin{tabular}{l | cc | cc | c}
        \toprule
        \multirow{2}{*}{Variant} & \multicolumn{2}{c|}{Spatial Supervision} & \multicolumn{2}{c|}{Spatial Predictor} & \multirow{2}{*}{Acc} \\
        \cmidrule(lr){2-3} \cmidrule(lr){4-5}
        & Position & Scale & PAttn & FAttn &  \\
        \midrule
        1 & \checkmark &            &            & \checkmark & 63.9 \\
        2 &            & \checkmark &            & \checkmark & 63.0 \\
        3 & \checkmark & \checkmark &            &            & 64.3 \\
        4 & \checkmark & \checkmark & \checkmark &            & 63.7 \\
        \rowcolor{black!4}
        5 (Ours)  & \checkmark & \checkmark &            & \checkmark & \textbf{64.8} \\
        \bottomrule
        \end{tabular}
        }
    \end{minipage}\vspace{-4mm}
\end{table*}


\noindent \textbf{The benefits of SP in spatial reasoning transfer to downstream tasks via coherent part-to-whole reconstruction.}
We demonstrate the effectiveness of SP for spatial reasoning using the jigsaw understanding task. As shown in \textbf{Tab.~\ref{tab:tab2}c}, SP consistently improves permutation prediction accuracy across all SSL backbones, with large gains for MoCo v3 (69.87 $\rightarrow$ 90.24) and MAE (77.95 $\rightarrow$ 98.58). This indicates that SP significantly strengthens the ability to infer global spatial configurations from disjoint local patches, a capability that is limited in standard SSL models.
This improved spatial reasoning further benefits downstream tasks such as recognition on reconstructed images, where patches are rearranged according to predicted permutations. As shown in \textbf{Tab.~\ref{tab:tab2}d}, SP consistently improves recognition accuracy across all models, including MAE (39.19 $\rightarrow$ 48.21) and MoCo v3 (57.09 $\rightarrow$ 64.23).
These results suggest that the recovered spatial structures are both geometrically accurate and semantically meaningful. By enforcing spatial consistency during pre-training, SP encourages representations that preserve coherent part-to-whole relationships under spatial reorganization.



\subsection{Ablation Studies Reveal Key Design Choices in SP.}

We conduct ablation studies on the SP design using a MoCo v3 (ViT-Tiny) backbone pre-trained on CIFAR-100 for in-domain image classification. Top-1 accuracy is reported in \textbf{Tab.~\ref{tab:ablation_checkmark_optimized}}.

\noindent \textbf{Joint supervision of position and scale is critical for visual representation learning.}
In Columns 1 and 2 of \textbf{Tab.~\ref{tab:ablation_checkmark_optimized}}, position-only L2 regression (Variant 1) achieves 63.9\% accuracy, while scale-only supervision (Variant 2) yields 63.0\%. However, both are outperformed by our full SP (Variant 5), which jointly models position and scale. This indicates that the two objectives provide complementary supervision, and that joint regression better captures spatial structure in a unified representation.

\noindent \textbf{Projection-free cross-attention is effective for spatial representation learning.}
In Columns 3 and 4 of \textbf{Tab.~\ref{tab:ablation_checkmark_optimized}}, we evaluate the role of cross-attention and its parameterization. Variant 3 removes cross-attention and directly concatenates patch tokens from the two views $I_r$ and $I_t$, followed by a 2-layer MLP for prediction, achieving 64.3\% accuracy. This simple concatenation of feature tokens already shows strong performance, suggesting that simple aggregation provides a strong baseline for spatial reasoning.
Next, Variant 4 introduces linear projections for queries, keys, and values in cross-attention. This degrades performance to 63.7\%, compared to 64.8\% for our full SP. This suggests that additional parameterization may dilute the spatial supervision signal. In contrast, preserving raw token interactions while enforcing structured cross-view attention is more effective for learning spatial relationships.

\section{Discussion}
\label{sec: discussion}

We present Spatial Prediction (SP), a spatially aware pretext task that explicitly models relative position and scale between two local views from the same image during SSL. SP serves as a simple plug-in for existing SSL frameworks, without modifying the underlying architectures or incurring additional inference cost.
We introduce a comprehensive evaluation suite spanning 7 downstream tasks, 11 datasets, 6 SSL models, 2 backbones, and 7 metrics. Among these, we further propose two spatial reasoning benchmarks: spatial prediction and jigsaw understanding. Empirical results indicate that existing SSL models exhibit limited spatial reasoning ability. In contrast, SP substantially improves spatial reasoning, while also enhancing semantic representation quality, robustness to corruptions and occlusions, and reducing texture bias in favor of shape-based representations.
Moreover, SP-learned representations are spatially grounded and transfer effectively to semantic segmentation and fine-grained classification. They also capture 3D geometric structure, yielding improvements on depth estimation. Our benchmarks further show that SP’s inductive bias enables recovery of structured spatial layouts from disorganized patches, whereas implicit positional embeddings in standard SSL are insufficient for spatial reasoning.
Overall, SP provides a simple yet effective mechanism for bridging semantic and geometric learning in self-supervised representations, and motivates future work on spatial, temporal, and physical reasoning in visual foundation models.

Despite its strong spatial reasoning performance, SP is currently limited to 2D space. Extending it to 3D or temporal constraints is an important direction for future work. In addition, while SP is designed as a plug-in regularizer, its performance may benefit from more advanced local view sampling strategies. We hope that stronger spatial reasoning capabilities will enable more capable vision foundation models for robotics, assistive technologies, and embodied AI.

\bibliographystyle{unsrt}
\bibliography{main}







\appendix
\newpage

\renewcommand{\thesection}{S\arabic{section}}
\renewcommand{\thefigure}{S\arabic{figure}}
\renewcommand{\thetable}{S\arabic{table}}
\setcounter{figure}{0}
\setcounter{section}{0}
\setcounter{table}{0}

\section*{Supplementary Materials}


\begin{figure}[h]
    \centering
    \includegraphics[width=1\linewidth]{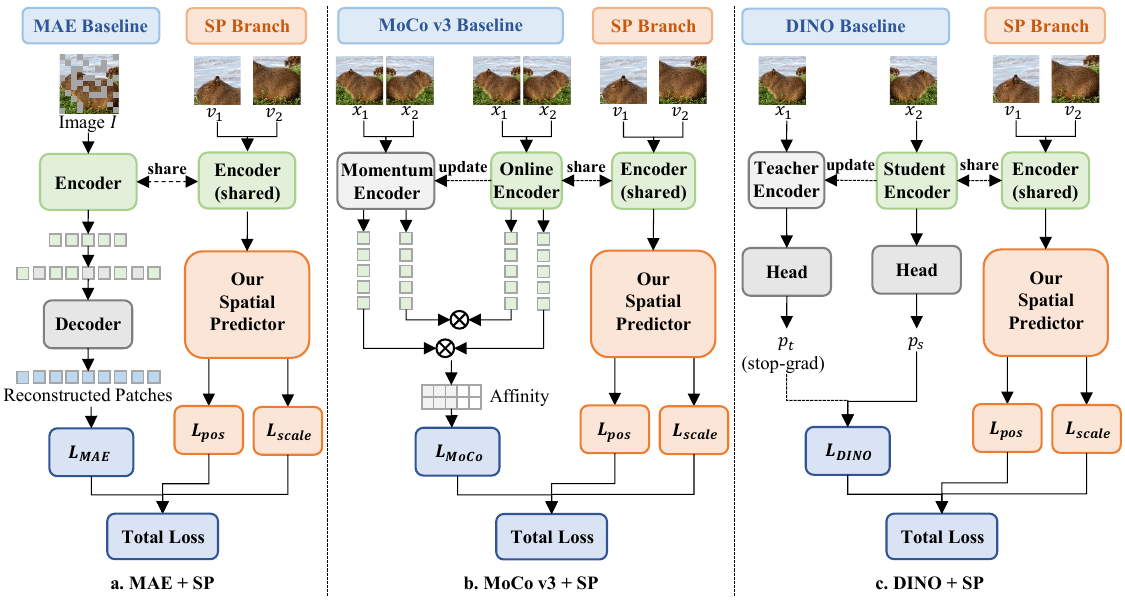}
    \caption{\textbf{Detailed architecture of Spatial Prediction (SP) integration across diverse SSL frameworks.} As a plug-in pretext task, SP is seamlessly incorporated into MAE, MoCo v3, and DINO. 
    (a) \textbf{MAE + SP:} To integrate SP into MAE, we share the base MAE encoder $\psi(\cdot)$. During a forward pass, the baseline MAE computes the reconstruction loss $\mathcal{L}_{\text{MAE}}$ using the masked image $I$. Concurrently, the SP branch takes two separate augmented views $v_1$ and $v_2$ as input. The shared encoder $\psi(\cdot)$ processes $v_1$ and $v_2$ to output the corresponding patch tokens $\mathbf{Z}_{v1}$ and $\mathbf{Z}_{v2}$. The Spatial Predictor takes these tokens to compute the spatial regression loss $\mathcal{L}_{\text{SP}} = \mathcal{L}_{pos} + \mathcal{L}_{scale}$. 
    (b) \textbf{MoCo v3 + SP:} When incorporating SP into MoCo v3, we specifically share the \textbf{online encoder} $\psi_{\text{online}}(\cdot)$ as the feature extractor for the SP branch. While the baseline computes the MoCo loss $\mathcal{L}_{\text{MoCo}}$ using the similarities between the online and momentum projections of $x_1$ and $x_2$, the SP branch processes its own views $v_1$ and $v_2$ through the shared $\psi_{\text{online}}(\cdot)$. This design forces the online encoder to not only achieve invariance to heavy augmentations (via $\mathcal{L}_{\text{MoCo}}$) but also maintain an explicit understanding of spatial equivariance (via $\mathcal{L}_{\text{SP}}$). The gradients from both $\mathcal{L}_{\text{MoCo}}$ and $\mathcal{L}_{\text{SP}}$ are backpropagated through the online encoder. 
    (c) \textbf{DINO + SP:} For integration with DINO, the SP branch specifically shares the \textbf{student encoder} $\psi_{\text{student}}(\cdot)$. The views $v_1$ and $v_2$ are fed into $\psi_{\text{student}}(\cdot)$ to extract the class tokens $\mathbf{z}_{v1}$ and $\mathbf{z}_{v2}$ (and/or patch tokens), which are then passed to the Spatial Predictor. Crucially, the gradients from the SP auxiliary loss $\mathcal{L}_{\text{SP}}$ only update the student encoder, while the teacher encoder remains updated solely via exponential moving average (EMA).}
    \label{fig:figS1}
\end{figure}

\clearpage
\begin{table}[h]
\centering
\caption{\textbf{Pre-training hyperparameters for 100-epoch evaluation.} We maintain identical optimization settings between each baseline and its $+SP$ variant to ensure a fair comparison. All models are trained on ImageNet-1K with a total batch size of 256. We borrow code from sololearn~\cite{JMLR:v23:21-1155}, lightly\cite{Susmelj_Lightly}, and DINO\cite{caron2021dino, oquab2024dinov2, si2025dinov3}}
\label{tab:hyperparams_final}
\resizebox{\textwidth}{!}{%
\begin{tabular}{lcc ccc ccc}
\toprule
\multirow{2}{*}{Hyperparameters} & \multicolumn{2}{c}{MAE (ViT-B)} & & \multicolumn{2}{c}{MoCo v3 (ViT-S)} & & \multicolumn{2}{c}{DINO (ViT-S)} \\ 
\cmidrule{2-3} \cmidrule{5-6} \cmidrule{8-9}
& Baseline & +SP & & Baseline & +SP & & Baseline & +SP \\
\midrule
\textit{Optimization} & & & & & & & & \\
Optimizer & \multicolumn{2}{c}{AdamW} & & \multicolumn{2}{c}{AdamW} & & \multicolumn{2}{c}{AdamW} \\
Base Learning Rate & \multicolumn{2}{c}{1.5e-4} & & \multicolumn{2}{c}{1.5e-4} & & \multicolumn{2}{c}{5e-4} \\
Weight Decay & \multicolumn{2}{c}{0.05} & & \multicolumn{2}{c}{0.1} & & \multicolumn{2}{c}{0.04 $\rightarrow$ 0.4} \\
Optimizer Momentum & \multicolumn{2}{c}{$\beta_{1,2}$=(0.9, 0.95)} & & \multicolumn{2}{c}{$\beta_{1,2}$=(0.9, 0.999)} & & \multicolumn{2}{c}{$\beta_{1,2}$=(0.9, 0.999)} \\
\midrule
\textit{Training Schedule} & & & & & & & & \\
Total Epochs & \multicolumn{2}{c}{100} & & \multicolumn{2}{c}{100} & & \multicolumn{2}{c}{100} \\
Batch Size & \multicolumn{2}{c}{256} & & \multicolumn{2}{c}{256} & & \multicolumn{2}{c}{256} \\
Learning Rate Schedule & \multicolumn{8}{c}{Cosine decay} \\
Warmup Epochs & \multicolumn{2}{c}{10} & & \multicolumn{2}{c}{10} & & \multicolumn{2}{c}{10} \\
\midrule
\textit{Method-Specific} & & & & & & & & \\
Masking Ratio & 75\% & 75\% & & N/A & N/A & & N/A & N/A \\
EMA Momentum & N/A & N/A & & \multicolumn{2}{c}{0.99 $\rightarrow$ 1.0} & & \multicolumn{2}{c}{0.996 $\rightarrow$ 1.0} \\
Temperature ($\tau$) & N/A & N/A & & \multicolumn{2}{c}{0.2} & & \multicolumn{2}{c}{0.04 $\rightarrow$ 0.07} \\
Multi-crop Scale & N/A & N/A & & N/A & N/A & & \multicolumn{2}{c}{(0.4, 1.0) / (0.05, 0.4)} \\
\midrule
\rowcolor[HTML]{F2F2F2} 
SP Loss Weight ($\lambda$) & -- & 0.1 & & -- & 0.1 & & -- & 0.1 \\
\bottomrule
\end{tabular}%
}
\end{table}

\clearpage
\begin{figure}[h]
    \centering
    \includegraphics[width=1.0\textwidth]{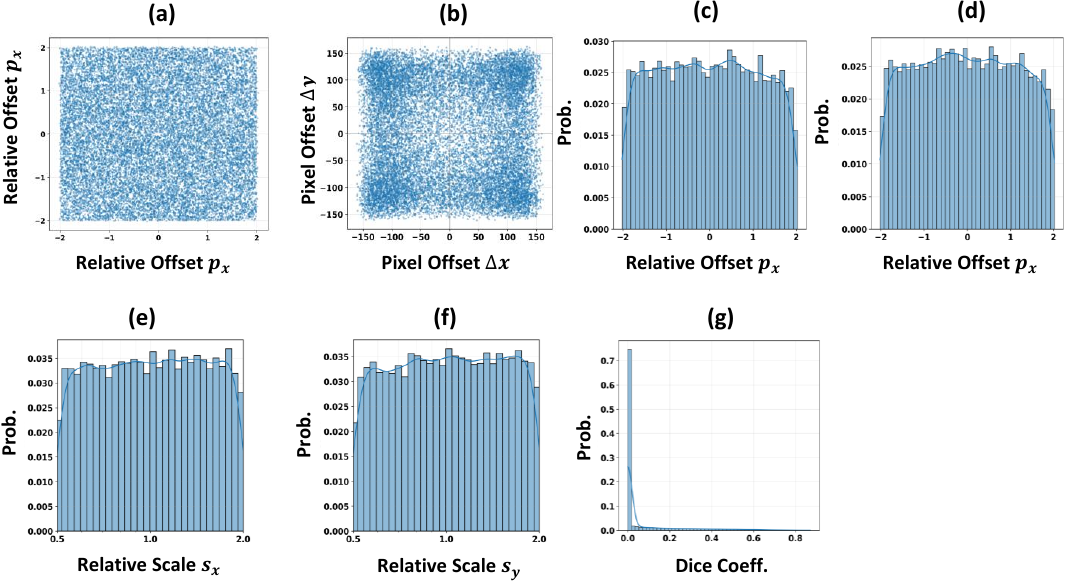}
    \caption{\textbf{Statistical validation of the spatial view sampling strategy.}
    Our sampling strategy is designed with three objectives:
    (i) generating semantically meaningful local views~\cite{caron2021dino, zhou2022ibot, assran2023ijepa},
    (ii) inducing a broad and approximately uniform distribution over spatial prediction targets (relative position and scale),
    and (iii) avoiding degenerate pairs with excessive overlap.
    To verify that our rejection sampling accurately reproduces the intended distributions while respecting the image boundaries, we simulated the generation of $N = 20{,}000$ pairs of local views. 
        \textbf{(a)} Joint distribution of relative offsets $(p_x, p_y)$. \textbf{(b)} Scatter plot of raw pixel offsets $(\Delta x, \Delta y)$. \textbf{(c, d)} Marginal distributions of $p_x$ and $p_y$. \textbf{(e, f)} Log-uniform distributions of relative scales $s_x$ and $s_y$. \textbf{(g)} Distribution of pairwise view Dice overlap.
        The sampled relative offsets and log-scale ratios approximately follow the intended uniform distributions in (a--f), indicating that the rejection sampling procedure preserves the target distributions despite finite image boundary constraints.
        Furthermore, the Dice overlap distribution in (g) is concentrated near zero, suggesting that the sampler effectively suppresses highly overlapping and potentially redundant view pairs while maintaining spatial diversity.
        }
    \label{fig:figS2}
\end{figure}

\clearpage
\begin{table}[h]
\centering
\caption{\textbf{Ablation on spatial loss weight $\lambda$.} We report the CIFAR-100 linear probing (LP) top-1 accuracy. The experiments are conducted using the MoCoV3 (ViT-S) backbone pre-trained for 800 epochs. Setting $\lambda_p=\lambda_s=0.1$ yields the best balance between semantic representation and spatial reasoning.}
\label{tab:ablation_lambda}
\begin{tabular}{lc}
\toprule
\textbf{Spatial Loss Weight ($\lambda_p=\lambda_s$)} & \textbf{CIFAR-100 LP Acc (\%)} \\
\midrule
$\lambda_p=\lambda_s=0.05$ & 63.5 \\
\rowcolor{black!4}
$\lambda_p=\lambda_s=0.1$ (Ours) & \textbf{64.8} \\
$\lambda_p=\lambda_s=0.5$ & 62.7 \\
\bottomrule
\end{tabular}
\end{table}

\newpage
\begin{figure}[h]
    \centering
    \includegraphics[width=1\linewidth]{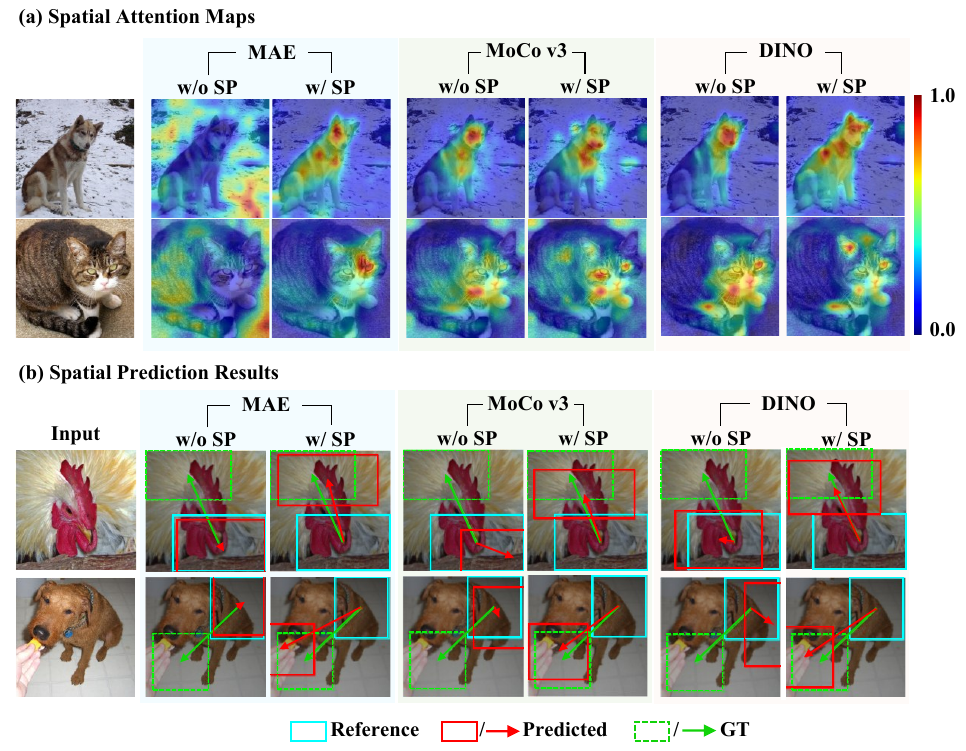}
    \caption{\textbf{Additional visualization examples for MoCo v3 with and without SP.}
    We provide additional qualitative examples following the same design convention as \textbf{Fig.\ref{fig:fig3}} in the main text.
    }
    \label{fig:figS3}
\end{figure}

\clearpage
\begin{figure}[H]
    \centering
    \includegraphics[width=1\linewidth]{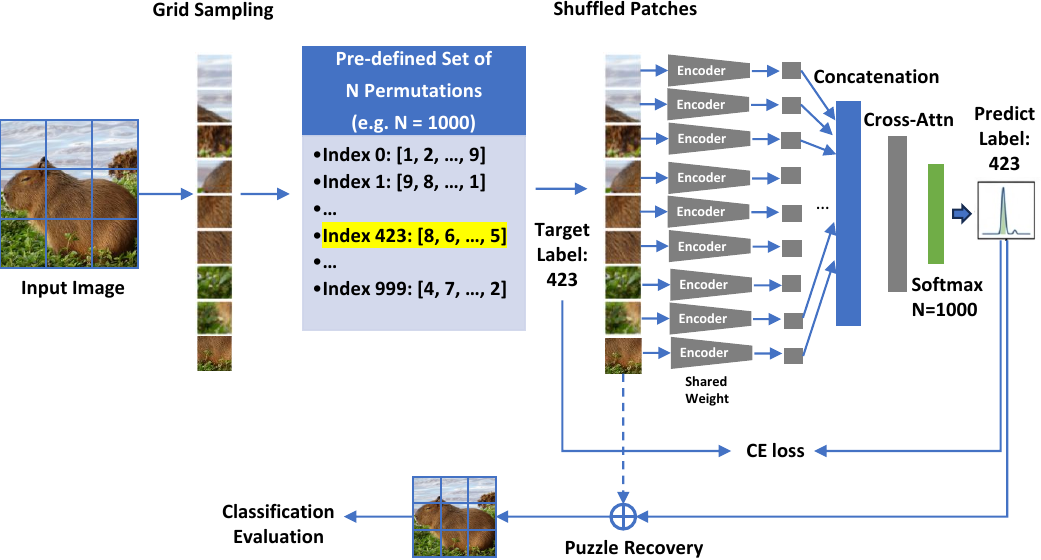}
    \caption{\textbf{Overview of the Jigsaw Understanding Evaluation.} To assess the joint modeling of spatial and semantic information, we construct a jigsaw reconstruction task following the formulation of ~\cite{noroozi2016jigsaw}. Given an input image, nine local patches are sampled and shuffled according to a target index from a predefined set of $N=1,000$ permutations. These permutations are selected to maximize pairwise Hamming distance, thereby mitigating shortcut learning. Each patch is independently processed through a shared-weight encoder to extract semantic features. In our design, a Cross-Attention mechanism is employed to enable each patch query to reason about its relative position conditioned on the global context of all other patches. The head is trained using Cross-Entropy (CE) loss to predict the permutation index. During evaluation, we perform puzzle recovery based on the predicted index and feed the reconstructed image back into the frozen pre-trained backbone with its linear probe to obtain classification performance. }
    \label{fig:figS4}
\end{figure}

\newpage

\end{document}